\documentclass{article}

\usepackage{arxiv}

\usepackage[utf8]{inputenc} 
\usepackage[T1]{fontenc}    
\usepackage{hyperref}       
\usepackage{url}            
\usepackage{booktabs}       
\usepackage{amsfonts}       
\usepackage{nicefrac}       
\usepackage{microtype}      
\usepackage{xcolor}         
\usepackage{multicol}

\usepackage{graphicx} 
\usepackage[utf8]{inputenc}
\usepackage{float}
\usepackage{subcaption}
\usepackage{natbib}
\usepackage{authblk}
\usepackage{url}
\usepackage[acronym,nogroupskip,nopostdot]{glossaries}
\usepackage{booktabs}
\usepackage{xltabular}
\usepackage{multirow, multicol, makecell, booktabs}
\usepackage{amsmath}
\usepackage{amsfonts}
\usepackage{optidef}
\usepackage{subfiles}
\usepackage{nameref}
\usepackage{hyperref}
\usepackage{adjustbox}
\usepackage{indentfirst}
\usepackage{array,ragged2e}
\usepackage{pdflscape}
\usepackage{longtable}
\hypersetup{
     colorlinks = true,
     citecolor  = blue,
     linkcolor = blue,
     urlcolor = blue
}
\usepackage{cleveref}
\usepackage{geometry}
\graphicspath{ {./images/} }

\usepackage{setspace}

\setcitestyle{authoryear,open={(},close={)}} 

\usepackage{tikz}
\usetikzlibrary{shapes,backgrounds}

\tikzstyle{block} = [rectangle, draw, fill=blue!20, text width=6em, text centered, rounded corners, minimum height=4em]
\tikzstyle{line} = [draw, -latex']

\title{Beyond Algorithmic Fairness: A Guide to Develop and Deploy Ethical AI-Enabled Decision-Support Tools}

\author{
  Rosemarie Santa Gonz\'alez\\
  H. Milton Stewart School of Industrial and Systems Engineering\\
  Georgia Institute of Technology\\
  \texttt{rosemarie.santa@gatech.edu} \\
  \and
  Ryan Piansky\\
  School of Electrical and Computer Engineering\\
  Georgia Institute of Technology \\
  \texttt{rpiansky3@gatech.edu} \\
  \and
  Sue M Bae\\
  School of Public Policy \\
  Georgia Institute of Technology \\
  \texttt{sbae99@gatech.edu} \\
  \and
  Justin Biddle \\
  School of Public Policy \\
  Georgia Institute of Technology \\
  \texttt{justin.biddle@pubpolicy.gatech.edu} \\
  \and
  Daniel Molzahn \\
  School of Electrical and Computer Engineering\\
  Georgia Institute of Technology \\
  \texttt{molzahn@gatech.edu}
}

\begin{document}

\date{ }

\maketitle

\begin{abstract}
  The integration of artificial intelligence (AI) and optimization hold substantial promise for improving the efficiency, reliability, and resilience of engineered systems. Due to the networked nature of many engineered systems, ethically deploying methodologies at this intersection poses challenges that are distinct from other AI settings, thus motivating the development of ethical guidelines tailored to AI-enabled optimization. This paper highlights the need to go beyond fairness-driven algorithms to systematically address ethical decisions spanning the stages of modeling, data curation, results analysis, and implementation of optimization-based decision support tools. Accordingly, this paper identifies ethical considerations required when deploying algorithms at the intersection of AI and optimization via case studies in power systems as well as supply chain and logistics. Rather than providing a prescriptive set of rules, this paper aims to foster reflection and awareness among researchers and encourage consideration of ethical implications at every step of the decision-making process. 
\end{abstract}

\section{Introduction}

The integration of artificial intelligence (AI) and optimization is transforming the landscape of engineered systems, offering unprecedented opportunities to enhance efficiency, reliability, and resilience across domains \citep{palle2023evolutionary} such as power systems \citep{thirunavukkarasu2023comprehensive}, supply chains, and logistics \citep{joel2024leveraging}. As these networked systems become more dependent on AI-enabled decision support tools, the ethical challenges associated with their deployment grow more complex \citep{whittlestone2022ai}. Traditional ethical concerns in AI—such as fairness, accountability, and transparency—take on new dimensions when applied to systems characterized by complex networks and optimization processes, where decisions have far-reaching societal impacts \citep{jobin2019global}.

Governments and organizations worldwide have responded to these ethical concerns by introducing frameworks and regulations aimed at ensuring trustworthy AI \citep{harrison2022cultivating, weaver4federal, aoki2024trustworthy, madhavan2020toward}. Initiatives like the European Union's AI Act \citep{eu_ai_act_trilogue_2024} and the Biden-Harris administration's AI Bill of Rights \citep{biden_executive_order_14110} aim to safeguard fairness, transparency, and accountability in AI systems \citep{ai_bill_of_rights, OECD2020,radu2021steering}. \textcolor{black}{However, while these efforts have made significant progress, much of the existing literature focuses predominantly on algorithmic fairness, often neglecting the broader implications of AI systems in optimization-based decision-making}\citep{fazelpour2020algorithmic}. The challenge remains to develop ethical guidelines that address not only fairness-driven algorithms but also the wider array of ethical considerations present in the entire pipeline of AI-enabled optimization tools.

This paper argues that ethical concerns in AI-enabled optimization extend beyond fairness, particularly when applied to complex networked systems. In many cases, fairness alone is insufficient to address the ethical complexities that arise during data curation, model development, result interpretation, and tool implementation. For example, optimization problems in power systems or supply chain networks often involve trade-offs that require careful ethical consideration to avoid perpetuating inequities or creating unintended consequences. Thus, we emphasize the need for a comprehensive ethical guidelines tailored to AI-enabled optimization of engineered systems that encompasses not only fairness but also transparency, accountability, inclusivity, and long-term societal impact.

Through case studies in power systems and supply chains, this paper examines how ethical considerations manifest at different stages of the AI-enabled optimization process. The first case study examines a collaborative decentralized cold supply chain designed to support underrepresented minority communities, while the second focuses on optimizing power grid infrastructure to mitigate wildfire risks. These cases highlight common ethical challenges—such as the handling of sensitive data, the transparency of modeling choices, and the interpretation of results—that are central to the responsible deployment of AI technologies. Rather than providing a prescriptive set of rules, this paper aims to foster reflection and awareness among researchers and practitioners, encouraging them to consider ethical implications at every step of the decision-making process. By drawing lessons from the case studies, we seek to demonstrate that AI-enabled decision support tools must go beyond optimizing for efficiency and fairness alone. To build truly responsible AI systems, we must also ensure they are transparent, accountable, and inclusive, and that they align with broader societal values. This paper calls for the development of ethical guidelines tailored to AI-enabled optimization, addressing not only fairness but also the ethical challenges unique to networked systems. Such guidelines will help safeguard the responsible and equitable deployment of AI technologies across various sectors, ensuring that they contribute positively to both communities and the environment.

The remainder of this paper is structured as follows: Section \ref{Sec:EAI} sets the stage for ethical considerations in AI-enabled optimization and presents the optimization process pipeline. Section \ref{Sec:Cases} presents the two case studies---one in supply chain logistics and the other in electric power systems---to explore the ethical challenges of AI-enabled optimization. Section \ref{Sec:Discussion} discusses the key ethical considerations across the stages of data collection, modeling, results interpretation, and implementation, drawing lessons from both case studies. Finally, Section \ref{Sec:Con} synthesizes the findings and calls for the development of comprehensive ethical frameworks for AI-enabled decision support tools.

\section{Ethical Considerations in AI-Enabled Optimization} \label{Sec:EAI}
AI-enabled optimization has the potential to enhance decision-making across diverse domains, but this advancement must be underpinned by ethical principles to prevent harm. Researchers must acknowledge that their models operate in complex, real-world environments where ethical risks, such as bias and inequity, arise at every stage—from data gathering to implementation \citep{johnson2023ethical}. To guide AI development responsibly, ethical principles must be integrated into the entire process, ensuring transparency, fairness, and societal well-being. Adoption of such approaches would mitigate risks such as unequal resource distribution and the amplification of bias in AI-driven systems.

\subsection{Optimization Models}
Optimization models are central to decision support tools that seek the best solution from a set of feasible options, typically with the goal of maximizing efficiency or improving outcomes \citep{taha2013operations}. AI enhances optimization by handling complex, dynamic systems and large datasets more effectively than traditional methods, uncovering patterns and adapting to changes \citep{bengio2021machine}. The interested reader is referred to \citep{van2024ai4opt} to learn more about the intersection of optimization and AI. However, the use of AI introduces new challenges, especially regarding fairness and bias. Researchers must consider the ethical implications of how models are constructed and how decisions are impacted by biased or incomplete data. For instance, algorithmic fairness plays a critical role in ensuring that AI-optimized models do not favor certain groups or reinforce systemic biases. It is essential to incorporate fairness criteria into the model design phase, ensuring that the solutions generated align with ethical principles and social justice goals. AI-driven optimization systems must be designed to promote equitable outcomes across diverse demographic groups \citep{veale2018fairness}. To accomplish this, ethical principles and fairness must be intertwined in the process as not doing so could lead to unintended negative consequences.

\subsection{Optimization Process}
\begin{figure}[ht!]
    \centering
    \includegraphics[width=.65\linewidth]{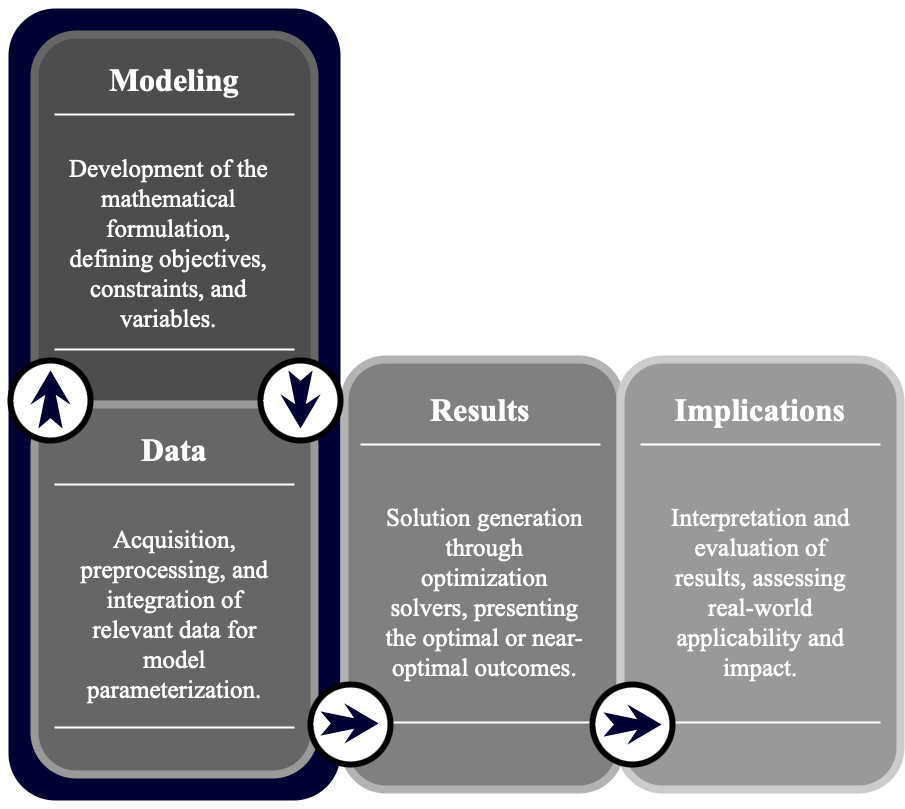}
    \caption{Four Stages of an Optimization Pipeline, from Data Collection, Modeling, Results, and Implications and Recommendations.}
    \label{fig: Optsteps}
\end{figure}

The optimization process is iterative, requiring continuous interaction between data, models, and solvers. At each stage, ethical considerations arise that can affect fairness and equity in decision-making. In Figure \ref{fig: Optsteps}, we depict the key stages of the optimization process and highlight where ethical considerations and fairness should be integrated. Throughout this process, researchers must engage with ethical principles that emphasize fairness and transparency, ensuring that AI-driven decision support tools are developed and deployed responsibly.

The process begins with data gathering, cleaning, and preparation, which can often introduce biases. If the data is unrepresentative or skewed, it can lead to biased results, reinforcing inequitable outcomes. Ethical data handling practices must be followed to minimize these risks. Additionally, preprocessing techniques that account for fairness can help in reducing biases during this stage. 
The data challenge is not unique to the optimization community; previous studies have shown similar issues in algorithms and machine learning. For instance, automation of data processing has been shown to impact machine learning algorithms \citep{guha2024automated}, while other researchers have demonstrated how fairness can be affected by the transformation of training data \citep{valentim2019impact}, and still others have highlighted how data preparation itself can lead to unfairness in AI systems \citep{khodadadian2021impact}.
When using solvers, whether commercial or open-source, their solutions should not be assumed fair without examination. Solvers typically identify one optimal solution, overlooking symmetric alternatives that lead to the same objective value. As shown in the study by \citet{st2022adaptation}, enumerating feasible solutions can impact the fairness of the final objective, underscoring the need to consider and examine all potential outcomes.
Finally, interpreting the results of the optimization model requires careful ethical evaluation. \citet{zhou2020towards} demonstrate that fair decision-making requires a deep contextual understanding, and AI explanations are essential in identifying the key variables driving unfair outcomes, highlighting the need for careful interpretation of results. Even though a solution may be mathematically optimal, it may not align with societal or ethical values, particularly when dealing with resource allocation or public services. After obtaining results, researchers must interpret them in a way that considers fairness and equity. Post-processing methods can be employed to adjust results and ensure that decision-making processes reflect ethical considerations. Throughout these stages, transparency and fairness should guide how researchers apply AI to optimization, ensuring that outcomes serve society equitably. Although this study focuses on optimization, these principles are not confined to this domain; the insights derived here are also relevant to algorithms and other AI-enabled decision support tools, where transparency and fairness remain crucial to ensuring ethical and equitable outcomes.

\section{Case Studies in AI-Enabled Optimization} \label{Sec:Cases}
To illustrate practical applications of various ethical considerations, we present two case studies: one in supply chain/logistics and the other on electric power systems. Both case studies involve complex, networked systems where decisions have far-reaching implications across interconnected components. Each case demonstrates how the principles of fairness, transparency, and responsible data handling influence AI-enabled optimization. These examples follow the structure of the optimization process outlined in  Figure \ref{fig: Optsteps}, showcasing the ethical challenges and decisions made at each stage within these networked systems.

\subsection{Case Study 1: Ethical Challenges in Collaborative Decentralized Cold Food Supply Chains}

The COVID-19 pandemic exposed vulnerabilities in food supply chains \citep{aday2020impact}, particularly in rural areas \citep{phillipson2020covid} where both farmers and end consumers from underrepresented minority communities faced significant barriers. Smallholder farmers, lacking the infrastructure and resources available to larger operations, were disproportionately affected by logistical disruptions, labor shortages, and rising costs \citep{hammond2022perceived}. Cold supply chains—critical for transporting temperature-sensitive goods like food—require expensive infrastructure, such as refrigerated storage and transportation \citep{wang2021cold}, which smallholder farmers struggle to access. This creates unreliable food supplies for consumers in these communities, who also experience heightened food insecurity. 

This case study focuses on a collaborative decentralized cold supply chain in rural Wisconsin, designed to empower smallholder farmers from underrepresented minority communities by pooling resources to serve their communities and participate in traditional markets. The goal is to ensure both farmers and consumers benefit from more reliable food supply systems. However, ethical concerns arise regarding fairness in resource allocation and ensuring that the benefits of collaboration are shared equitably among all participants, especially those with fewer resources.

Key challenges include ensuring that decision support tools reflect the needs of underrepresented minority farmers and their communities. Ethical decision-making must prioritize fairness in access to shared resources, such as transportation and storage, while considering the broader impact on food security for consumers in under-served areas. The goal is not just efficiency but also ensuring equitable benefits for both farmers and consumers, especially those from underrepresented groups.

\subsubsection{Ethical Considerations in Collaborative Decentralized Cold Supply Chains}
In developing decision support tools for collaborative decentralized cold supply chains, there are several ethical concerns to address at every stage, from data collection to modeling and final implementation. Each step must reflect the needs and priorities of underrepresented minority farmers and communities, ensuring fairness, transparency, and accountability.

\paragraph{Data}
In the context of the collaborative decentralized cold supply chain in rural Wisconsin, ethical data handling is crucial for building trust with underrepresented minority farmers and their communities. Due to historical marginalization, these groups often have concerns about data misuse or exploitation. Researchers must establish confidentiality measures, such as non-disclosure agreements, to protect data from competitors (\textit{i.e.,} larger-scale farmers) and ensure it is not publicly disclosed. Explicit consent for data use must be obtained, with transparency about how the data will be used, and researchers must respect decisions against sharing. Furthermore, collaboration in data validation and analysis, especially when developing metrics to measure benefits, ensures that the results reflect community priorities, such as the quantity and quality of food provided. This fosters transparency, trust, and continued participation from the communities involved.

\paragraph{Modeling}
Designing a model for a collaborative decentralized cold supply chain requires careful decision-making at every stage, from setting objectives to defining decision variables. The objective function is key, as focusing solely on cost minimization can exclude smaller, higher-cost farms, disadvantaging underrepresented minority farmers. Treating costs as constraints, while incorporating objectives like economic efficiency, social equity, and environmental sustainability, promotes a more balanced approach. Stakeholder input from underrepresented farmers ensures these objectives reflect community needs and account for trade-offs that may affect other marginalized groups. Model constraints should address the challenges of rural, decentralized systems, ensuring equitable resource access and incorporating social and environmental factors like emissions reduction and fair labor practices. The choice of decision variables also has significant implications for fairness—continuous variables allow for more nuanced and precise resource distribution, while binary variables simplify decisions but may fail to capture the complexities of diverse farm operations. Striking the right balance between the two is essential to ensure that the model fairly represents all stakeholders and supports ethical decision-making.

\paragraph{Results}
Ensuring ethical integrity in the results of AI and optimization tools requires deliberate effort, particularly in a collaborative decentralized food supply chain. The results directly affect various stakeholders, including farmers, pantries, and underrepresented groups, making transparency and explainability crucial. Researchers must ensure the model’s workings are clear and accessible, allowing stakeholders to understand how results are derived and identify any ethical concerns. Engaging all stakeholders in interpreting results helps align outcomes with community needs, while applying fairness metrics can detect and mitigate biases. For example, comparing resource allocations across farmers ensures equity, and adjustments to constraints or data can be made if disparities arise. Incorporating agricultural experts into the review process helps identify potential oversights, and continuous monitoring allows models to adapt to changing conditions, such as climate or infrastructure challenges. Regular impact assessments address both intended and unintended consequences, while cultural and contextual sensitivity ensures that results are communicated in an accessible and inclusive way, fostering fair outcomes for the entire supply chain.

\paragraph{Implications and Recommendations}
Optimization and AI models, especially in collaborative decentralized cold food supply chains, should be viewed as decision support tools that enhance, rather than replace, human judgment. Researchers must provide clear, comprehensive explanations to decision-makers to prevent misinterpretation or misuse, and involving experts and end-users throughout the modeling process helps anticipate potential complexities and misuses. Ethical considerations are crucial, as decisions in these supply chains directly affect the livelihoods of underrepresented minority farmers and the communities they serve. Poor logistics or inadequate refrigeration can lead to product spoilage or contamination, impacting consumer health. Therefore, decision support tools must prioritize fairness, inclusivity, and the well-being of all participants, as ethical modeling directly shapes the outcomes for interconnected communities, making it essential for developers to prioritize ethics in every aspect of their work.

\subsection{Case Study 2: Ethical Challenges in Power Systems’ Transmission Infrastructure}\label{PwrCase}

Power systems encompass various scales of infrastructure, from small-scale solar panels to continent-spanning transmission networks. The transmission grid is vital to modern life, enabling the delivery of power across large areas. Described as "the most complex machine ever made"~\citep{schewe2007grid}, the power grid faces increasing threats from natural disasters, especially as climate change worsens. Extreme weather can damage key infrastructure, such as substations and power lines, interrupting power delivery. Investments in infrastructure, such as line upgrades or battery placements, are essential but costly, making optimal planning critical. Additionally, understanding the impact on individual consumers and communities, particularly marginalized groups, is crucial as long-term infrastructure decisions are made. 

This case study focuses on an Optimal Transmission Switching (OTS) and line undergrounding investments problem under wildfire ignition risk~\citep{pollack2024equitably}. Technical details of the OTS formulation and methods used to allocate ignition risk are provided in the Appendix \ref{Apen:OTS}. Wildfires ignited by power lines, though rare, are often highly destructive~\citep{mitchell2013power}. To mitigate this risk, utilities often implement Public Safety Power Shutoff (PSPS) events, de-energizing lines to prevent fires but also causing power outages~\citep{huang2023review}. Power outages disproportionately affect marginalized communities~\citep{Ham_Lee_2022}, including those from PSPS events. The OTS-PSPS problem aims to minimize wildfire risk while ensuring equitable distribution of power to vulnerable communities, minimizing load shedding, and allocating investments like undergrounding power lines. With utilities like PG\&E committing billions to undergrounding power lines~\citep{PGE_undergrounding}, the question of how to fairly allocate resources is critical. The Biden-Harris administration’s Justice40 initiative aims to direct 40\% of the benefits from federally funded climate infrastructure investments to vulnerable communities~\citep{White_House_2023}. This case study evaluates how decision support tools can ensure these goals are met, balancing risk reduction with equitable outcomes. Further details on the data sources and integration methods used in this case study can be found in the Appendix \ref{Appen:PwrSys}.

\subsubsection{Ethical Considerations in Transmission Infrastructure}

In developing decision support tools for the power systems case study, we must consider ethical challenges across all four stages: data handling, modeling decisions, the presentation of results, and their broader implications for communities, especially marginalized populations.

\paragraph{Data}
The ethical curation of data in the power systems case study is crucial, particularly when dealing with demographic and vulnerability data from underrepresented communities. Data was sourced from multiple government databases, including the Justice40 designated census tracts, the Centers for Disease Control and Prevention (CDC) Social Vulnerability Index (SVI), and the United States Geological Survey (USGS) Wildland Fire Potential Index (WFPI). A primary ethical concern was ensuring that the data accurately represented vulnerable populations and that its use did not perpetuate historical inequities. The processes of correlating census tract data to the power grid and aggregating vulnerability metrics involved decisions that could potentially obscure the needs of smaller or less represented groups. For example, while the Justice40 initiative sets a national standard, local nuances may be lost when applying this standard across locally diverse regions like Texas. Therefore, it is important to redefine certain metrics based on regional averages to ensure equity at the local level. Details on the data sources and correlation methods are provided in Appendix \ref{Appen:PwrSys}.

\paragraph{Modeling}
The modeling process for the OTS problem also presents ethical considerations, especially in balancing wildfire risk mitigation with the equitable distribution of power. This model aimed to minimize the load shed while reducing the risk of wildfires, but key decisions about how to allocate resources—such as undergrounding power lines—have direct ethical implications. We considered multiple interpretations of the Justice40 initiative, including budget allocation and load shed reduction, and modeled impacts proportionally across demographic groups. This approach ensures that smaller or historically disadvantaged communities do not bear a disproportionate burden of power outages. However, the choice of objectives and constraints, such as whether to prioritize total load shed or proportional reductions, can significantly affect the fairness of the model’s outcomes. \cite{pollack2024equitably} discuss the formulation of these objectives and constraints.

\paragraph{Results}
When interpreting the results of the model, the ethical challenge lies in ensuring that outcomes are presented in a way that is clear, actionable, and fair. Results must be interpreted not only in terms of efficiency but also equity. For instance, while minimizing total system-wide load shed might seem optimal from a purely technical standpoint, this objective can overlook the specific needs of marginalized communities. This case study centers results on ensuring effective communication to policymakers and other stakeholders, allowing them to understand the trade-offs between minimizing total load shed and ensuring equity in power distribution. This study also highlights the need to ensure local communities and experts are presented with outcomes highlighting the impacts to different communities that may be most affected by wildfires and power outages. 

\paragraph{Implications and Recommendations}
The broader implications of this case study extend beyond the technical model itself. Infrastructure investments, such as undergrounding power lines, have long-lasting effects on the communities they serve. The decision support tools developed in this case study are designed to aid, not replace, human decision-making, empowering policymakers to weigh the ethical trade-offs of various strategies. An ethical framework allows decision-makers to have the autonomy to evaluate multiple options, each with its own set of risks and benefits. The results highlight the need for continuous monitoring and adaptation of models as new data becomes available, ensuring that infrastructure investments remain fair and equitable. Recommendations include understanding outcomes from reasonable interpretations of policies that ensure marginalized communities benefit proportionally from infrastructure investments, in line with the intent of initiatives like Justice40.

\section{Lessons Learned from the Case Studies} \label{Sec:Discussion}

In this section, we explore the ethical considerations that should be central to the development and deployment of AI-enabled decision support tools. By examining two distinct case studies, we aim to uncover common ethical challenges and considerations. The analysis is structured around key the stages of optimization: data, modeling, results, and implications. These categories allow us to systematically compare and contrast the ethical dimensions of each case study, providing insights into how researchers can address these challenges while going beyond fairness, addressing transparency, accountability, and long-term societal impacts. Table~\ref{tab:comparison} offers at a glance a quick comparison between the case studies in supply chain and power systems. 

\begin{table}[ht]
\centering
\caption{Ethical Considerations of Case Studies}
\label{tab:comparison}
\begin{tabularx}{\textwidth}{lXX}
\toprule
\textbf{Category} & \textbf{Supply Chain} & \textbf{Power Systems} \\
\midrule
\textbf{Data} & Sensitive data, sharing challenges & Incomplete datasets, multi-source correlation \\
\textbf{Modeling} & Fairness and equity in decision support & Focus on alternative solutions, ethical analysis \\
\textbf{Results} & Clear, interpretable results for stakeholders & Translation for policymakers \\
\textbf{Implications} & Tools supporting autonomy and transparency & Solutions verified by human decision-makers \\
\bottomrule
\end{tabularx}
\end{table}

With this comparison, we aim to generalize ethical considerations that researchers should follow to ensure responsible AI use in optimization, not only in terms of fairness but also broader ethical dimensions. This contributes to the ongoing dialogue on ethical AI and calls for the development of guidelines for AI-enabled decision support tools that go beyond fairness-driven algorithms. While existing ethical AI guidelines might address data privacy and transparency or encourage parity constraints \citep{correa2023worldwide, floridi2018ai4people, attard2023ethics}, optimization based algorithms require a deeper consideration of potential ethical impacts throughout the design process.

\subsection{Insights from the Case Studies}
Both case studies involve optimization processes within networked systems, but they highlight different ethical challenges. AI can be harnessed in both case studies to enhance decision support, but the ethical dimensions go beyond ensuring fairness. These include data sensitivity, transparency in modeling choices, and accountability for long-term impacts.

In the supply chain case, sensitive data from underrepresented minority communities brings added responsibility for privacy protection and ethical data handling. The power systems case study deals with multiple datasets, requiring complex integration. These distinct challenges emphasize the need for ethical data practices that address not only fairness but also transparency and accountability for data integrity. Modeling decisions differ but share a need for transparency and explainability. In supply chains, fairness is embedded through decision support tools, while in power systems, researchers must provide ethical analysis of multiple alternative solutions. Both approaches demonstrate the necessity of transparency in modeling choices and a broader ethical lens to ensure models are not only fair but also trustworthy and explainable. Lastly, the need for accessible and interpretable results is common to both cases, but the focus extends beyond fairness. Ensuring that results are actionable and transparent to all stakeholders, particularly underrepresented communities and policymakers, reinforces the need for responsible and ethical communication of AI-driven optimization results.

We next present insights derived from the two case studies for all four stages of the optimization process.

\paragraph{Data}
In the supply chain case study, the handling of sensitive data from underrepresented populations required a focus on trust, privacy, and long-term accountability. Researchers needed to ensure that the communities involved are protected from data misuse and that ethical principles such as transparency and respect for consent are maintained. In the power systems case, the need to merge datasets from various sources that are intended for different purposes presents challenges in ensuring data accuracy, consistency, and completeness. The ethical need for transparency in how data is combined and processed becomes crucial in avoiding biased or misleading results. Both cases highlight the importance of developing tools and methodologies that not only address fairness but also protect privacy, ensure transparency, and uphold accountability in how data is managed and used.

\paragraph{Modeling}
Modeling in both supply chain and power systems poses ethical challenges, with each requiring different approaches to fairness, transparency, and accountability. In supply chains, decision support tools must embed fairness and equity criteria to ensure just resource allocation. However, the ethical responsibility goes beyond fairness, requiring transparency in how objectives and constraints are set, and how these decisions impact different stakeholders. In power systems, modeling can involve providing multiple alternative solutions, with an emphasis on the ethical analysis of each. Researchers must ensure transparency in presenting these alternatives, giving decision-makers clear insights into the trade-offs involved. Both cases emphasize that models should not only ensure fair outcomes but also provide a transparent decision-making process that allows for accountability and stakeholder engagement.

\paragraph{Results}
Both case studies stress the importance of transparency and accountability in the presentation and interpretation of results. In the supply chain case, the focus on readable, interpretable results for underrepresented communities goes beyond fairness by ensuring that all stakeholders can actively participate in the decision-making process. Clear visualizations and transparent communication are essential to build trust and accountability. In the power systems case, researchers need to translate complex results into actionable insights for policymakers. The ethical challenge here involve ensuring that a range of results are not only fair but also transparent and understandable, enabling policymakers to make informed decisions. Third-party validation and continuous monitoring of results are critical in both cases, ensuring accountability for long-term societal impacts.

\paragraph{Implications and Recommendations}
Both case studies underscore the need for decision support tools to empower human decision-makers. In supply chains, these tools must incorporate fairness and equity criteria but also provide transparency in how decisions are supported. Decision-makers must be able to verify and evaluate solutions based on clear ethical guidelines that go beyond fairness and consider the broader societal and environmental implications. In power systems, tools should provide alternative solutions, allowing decision-makers to understand the ethical trade-offs between different options. This reinforces the need for AI tools to prioritize not just fairness, but also autonomy, transparency, and long-term accountability.

\subsection{Guidelines Informed by the Case Studies}
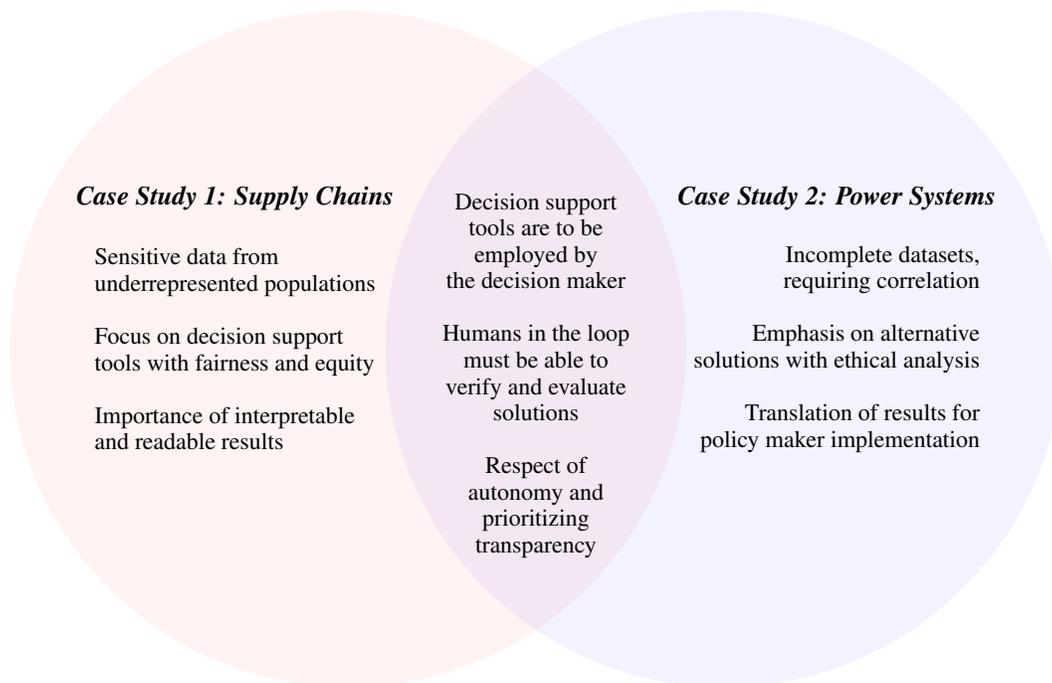
\begin{figure}[ht!]
    \centering
\begin{tikzpicture}
    \begin{scope}[fill opacity=0.05]
        \fill[red] (-2.5,0) circle (4.5);
        \fill[blue] (2.5,0) circle (4.5);
    \end{scope}
    \node at (-4,2) {\emph{\textbf{Case Study 1: Supply Chains}}};
    \node[align=left,font=\footnotesize] at (-4,0) {Sensitive data from\\underrepresented populations\\
    \\
         Focus on decision support\\tools with fairness and equity\\
         \\
         Importance of interpretable\\and readable results
    };
    \node at (4,2) {\emph{\textbf{Case Study 2: Power Systems}}};
    \node[align=right,font=\footnotesize] at (4,0) {Incomplete datasets,\\requiring correlation\\
    \\
        Emphasis on alternative\\solutions with ethical analysis\\
        \\
        Translation of results for\\policy maker implementation
    };
    \node[align=center,font=\footnotesize] at (0,0) {\phantom{Power Systems}\\ \phantom{Power Systems}\\Decision support\\ tools are to be\\ employed by\\the decision maker\\
    \\
        Humans in the loop\\ must be able to\\ verify and evaluate\\solutions\\
        \\
        Respect of\\ autonomy and\\ prioritizing\\ transparency
    };
\end{tikzpicture}
\caption{Summary of \textcolor{black}{E}thical Challenges \textcolor{black}{I}dentified \textcolor{black}{T}hrough \textcolor{black}{C}ase \textcolor{black}{S}tudies.}
    \label{fig:ScvPS}
\end{figure}

Despite differences between supply chain and power systems, common ethical themes emerge in the use of AI for optimization. See Figure \ref{fig:ScvPS} for a summary of the ethical challenges illustrated in  the case studies. Note that these common ethical themes also apply to other algorithm-based methods and applications. The case studies highlight the need to move beyond traditional algorithmic fairness and address broader concerns including transparency, accountability, inclusivity, and societal impact. Effective data management, clear modeling decisions, and interpretable results are crucial for ensuring that AI-enabled decision support tools are not only fair but also responsible and aligned with ethical standards.

For decision support tools to be aligned with ethical standards, they must prioritize transparency by documenting data sources, model assumptions, and decision-making processes in a way that is accessible to all stakeholders. Accountability mechanisms should be built into AI-enabled tools to ensure ongoing oversight and correction of biases or unintended consequences. Inclusivity must guide the development of models, actively engaging underrepresented communities to ensure equitable access to resources and decision-making power. Additionally, AI tools should be designed to empower, not replace, human decision-makers, enabling them to verify and evaluate solutions autonomously. Incorporating these principles and further developing guidelines and frameworks will help create AI-enabled decision support tools that are not only efficient but also socially responsible, ensuring that their long-term societal impacts are positive and equitable.

\section{Conclusion} \label{Sec:Con}
This paper demonstrates that ethical considerations in AI-enabled optimization must extend beyond fairness. While algorithmic fairness is essential for mitigating bias, it alone is insufficient to address the broader ethical challenges in the development and deployment of decision support tools. The existing literature on fairness primarily focuses on outcomes, often neglecting crucial elements such as the transparency of data and modeling processes, accountability mechanisms, and the long-term societal impacts of AI systems \citep{fazelpour2020algorithmic, goel2021importance}. These limitations create ethical blind spots where fairness-driven models may still perpetuate systemic inequities or fail to consider the broader consequences of decision making tools.

The two case studies discussed on supply chains and power systems illustrate that ethical challenges emerge at every stage of the optimization process---from data collection to modeling and result interpretation. The insights derrived from the case studies can be extrapolated to other applications and algorithms. Accordingly, we advocate for a more comprehensive ethical guidelines for AI-enabled decision support tools, ones that includes not only fairness but also transparency, accountability, inclusivity, and long-term societal impact. Ethical guidelines must be adopted and incorporated in the development and deployment of decision support tools. Such guidelines must prioritize transparency in documenting data sources, modeling decisions, and assumptions. Accountability mechanisms should ensure ongoing oversight and the ability to detect and address biases or unintended consequences. Furthermore, inclusivity must be at the heart of these systems, engaging underrepresented communities throughout the process to ensure their needs are reflected in the models. Finally, AI tools must enhance human autonomy by empowering decision-makers to verify and evaluate AI-driven solutions with full understanding of their ethical implications.

As AI continues to be integrated into complex systems like supply chains and power grids, developing ethical guidelines that go beyond fairness is critical to safeguarding equitable and responsible outcomes. These guidelines will ensure that AI technologies align with societal values, fostering transparency, equity, and long-term positive impacts on both communities and the environment.

\section{Acknowledgments}
This research was supported in part by the National Science Foundation (NSF) under award 2112533.

\bibliography{citation.bib}

\begin{thebibliography}{50}
\providecommand{\natexlab}[1]{#1}
\providecommand{\url}[1]{\texttt{#1}}
\expandafter\ifx\csname urlstyle\endcsname\relax
  \providecommand{\doi}[1]{doi: #1}\else
  \providecommand{\doi}{doi: \begingroup \urlstyle{rm}\Url}\fi

\bibitem[Aday and Aday(2020)]{aday2020impact}
S.~Aday and M.~S. Aday.
\newblock Impact of {COVID-19} on the food supply chain.
\newblock \emph{Food Quality and Safety}, 4\penalty0 (4):\penalty0 167--180, 2020.

\bibitem[Aoki et~al.(2024)Aoki, Tay, and Yarime]{aoki2024trustworthy}
N.~Aoki, M.~Tay, and M.~Yarime.
\newblock Trustworthy public sector {AI}: Research progress and future agendas.
\newblock \emph{Research Handbook on Public Management and Artificial Intelligence}, pages 263--276, 2024.

\bibitem[Attard-Frost et~al.(2023)Attard-Frost, De~los R{\'\i}os, and Walters]{attard2023ethics}
B.~Attard-Frost, A.~De~los R{\'\i}os, and D.~R. Walters.
\newblock The ethics of {AI} business practices: A review of 47 {AI} ethics guidelines.
\newblock \emph{AI and Ethics}, 3\penalty0 (2):\penalty0 389--406, 2023.

\bibitem[Awad et~al.(2021)Awad, Ndiaye, and Osman]{awad2021vehicle}
M.~Awad, M.~Ndiaye, and A.~Osman.
\newblock Vehicle routing in cold food supply chain logistics: A literature review.
\newblock \emph{The International Journal of Logistics Management}, 32\penalty0 (2):\penalty0 592--617, 2021.

\bibitem[Bengio et~al.(2021)Bengio, Lodi, and Prouvost]{bengio2021machine}
Y.~Bengio, A.~Lodi, and A.~Prouvost.
\newblock Machine learning for combinatorial optimization: A methodological tour d’horizon.
\newblock \emph{European Journal of Operational Research}, 290\penalty0 (2):\penalty0 405--421, 2021.

\bibitem[Biden(2021)]{biden_executive_order_14110}
J.~Biden.
\newblock Executive order 14110: Strengthening american leadership in artificial intelligence, 2021.
\newblock URL \url{https://www.federalregister.gov/documents/2021/06/11/2021-12499/strengthening-american-leadership-in-artificial-intelligence}.
\newblock Accessed: 2024-09-12.

\bibitem[Corr{\^e}a et~al.(2023)Corr{\^e}a, Galv{\~a}o, Santos, Del~Pino, Pinto, Barbosa, Massmann, Mambrini, Galv{\~a}o, Terem, et~al.]{correa2023worldwide}
N.~K. Corr{\^e}a, C.~Galv{\~a}o, J.~W. Santos, C.~Del~Pino, E.~P. Pinto, C.~Barbosa, D.~Massmann, R.~Mambrini, L.~Galv{\~a}o, E.~Terem, et~al.
\newblock Worldwide {AI} ethics: A review of 200 guidelines and recommendations for {AI} governance.
\newblock \emph{Patterns}, 4\penalty0 (10), 2023.

\bibitem[Fazelpour and Lipton(2020)]{fazelpour2020algorithmic}
S.~Fazelpour and Z.~C. Lipton.
\newblock Algorithmic fairness from a non-ideal perspective.
\newblock In \emph{Proceedings of the AAAI/ACM Conference on AI, Ethics, and Society}, pages 57--63, 2020.

\bibitem[Floridi et~al.(2018)Floridi, Cowls, Beltrametti, Chatila, Chazerand, Dignum, Luetge, Madelin, Pagallo, Rossi, et~al.]{floridi2018ai4people}
L.~Floridi, J.~Cowls, M.~Beltrametti, R.~Chatila, P.~Chazerand, V.~Dignum, C.~Luetge, R.~Madelin, U.~Pagallo, F.~Rossi, et~al.
\newblock {AI4People}---{An} ethical framework for a good {AI} society: Opportunities, risks, principles, and recommendations.
\newblock \emph{Minds and Machines}, 28:\penalty0 689--707, 2018.

\bibitem[Goel et~al.(2021)Goel, Amayuelas, Deshpande, and Sharma]{goel2021importance}
N.~Goel, A.~Amayuelas, A.~Deshpande, and A.~Sharma.
\newblock The importance of modeling data missingness in algorithmic fairness: A causal perspective.
\newblock In \emph{Proceedings of the AAAI Conference on Artificial Intelligence}, volume 35-9, pages 7564--7573, 2021.

\bibitem[Guajardo and Rönnqvist(2015)]{GUAJARDO2015147}
M.~Guajardo and M.~Rönnqvist.
\newblock Operations research models for coalition structure in collaborative logistics.
\newblock \emph{European Journal of Operational Research}, 240\penalty0 (1):\penalty0 147--159, 2015.

\bibitem[Guha et~al.(2024)Guha, Khan, Stoyanovich, and Schelter]{guha2024automated}
S.~Guha, F.~A. Khan, J.~Stoyanovich, and S.~Schelter.
\newblock Automated data cleaning can hurt fairness in machine learning-based decision making.
\newblock \emph{{\rm To appear in} IEEE Transactions on Knowledge and Data Engineering}, 2024.

\bibitem[Ham and Lee(2022)]{Ham_Lee_2022}
Y.~Ham and S.~Lee.
\newblock Behavior analysis of socially vulnerable households responding to planned power shutoffs.
\newblock \emph{Natural Hazards Center Mitigation Matters Grant Report}, 2022.
\newblock URL \url{https://hazards.colorado.edu/mitigation-matters-report/behavior-analysis-of-socially-vulnerable-households-responding-to-planned-power-shutoffs}.

\bibitem[Hammond et~al.(2022)Hammond, Siegal, Milner, Elimu, Vail, Cathala, Gatera, Karim, Lee, Douxchamps, et~al.]{hammond2022perceived}
J.~Hammond, K.~Siegal, D.~Milner, E.~Elimu, T.~Vail, P.~Cathala, A.~Gatera, A.~Karim, J.-E. Lee, S.~Douxchamps, et~al.
\newblock Perceived effects of {COVID-19} restrictions on smallholder farmers: Evidence from seven lower- and middle-income countries.
\newblock \emph{Agricultural Systems}, 198:\penalty0 103367, 2022.

\bibitem[Harrison and Luna-Reyes(2022)]{harrison2022cultivating}
T.~M. Harrison and L.~F. Luna-Reyes.
\newblock Cultivating trustworthy artificial intelligence in digital government.
\newblock \emph{Social Science Computer Review}, 40\penalty0 (2):\penalty0 494--511, 2022.

\bibitem[Huang et~al.(2023)Huang, Hu, Sang, Lucas, Wong, Wang, Hong, Yao, and Donde]{huang2023review}
C.~Huang, Q.~Hu, L.~Sang, D.~D. Lucas, R.~Wong, B.~Wang, W.~Hong, M.~Yao, and V.~Donde.
\newblock A review of public safety power shutoffs {(PSPS)} for wildfire mitigation: Policies, practices, models and data sources.
\newblock \emph{IEEE Transactions on Energy Markets, Policy and Regulation}, 1\penalty0 (3):\penalty0 187--197, 2023.

\bibitem[Jobin et~al.(2019)Jobin, Ienca, and Vayena]{jobin2019global}
A.~Jobin, M.~Ienca, and E.~Vayena.
\newblock The global landscape of ai ethics guidelines.
\newblock \emph{Nature Machine Intelligence}, 1\penalty0 (9):\penalty0 389--399, 2019.

\bibitem[Joel et~al.(2024)Joel, Oyewole, Odunaiya, and Soyombo]{joel2024leveraging}
O.~S. Joel, A.~T. Oyewole, O.~G. Odunaiya, and O.~T. Soyombo.
\newblock Leveraging artificial intelligence for enhanced supply chain optimization: A comprehensive review of current practices and future potentials.
\newblock \emph{International Journal of Management \& Entrepreneurship Research}, 6-3:\penalty0 707--721, 2024.

\bibitem[Johnson and Verdicchio(2023)]{johnson2023ethical}
D.~G. Johnson and M.~Verdicchio.
\newblock Ethical {AI} is not about {AI}.
\newblock \emph{Communications of the ACM}, 66\penalty0 (2):\penalty0 32--34, 2023.

\bibitem[Jung et~al.(2008)Jung, {Frank Chen}, and Jeong]{JUNG2008348}
H.~Jung, F.~{Frank Chen}, and B.~Jeong.
\newblock Decentralized supply chain planning framework for third party logistics partnership.
\newblock \emph{Computers \& Industrial Engineering}, 55\penalty0 (2):\penalty0 348--364, 2008.

\bibitem[Khodadadian et~al.(2021)Khodadadian, Ghassami, and Kiyavash]{khodadadian2021impact}
S.~Khodadadian, A.~Ghassami, and N.~Kiyavash.
\newblock Impact of data processing on fairness in supervised learning.
\newblock In \emph{2021 IEEE International Symposium on Information Theory (ISIT)}, pages 2643--2648. IEEE, 2021.

\bibitem[Kody et~al.(2022)Kody, Piansky, and Molzahn]{kody2022optimizing}
A.~Kody, R.~Piansky, and D.~K. Molzahn.
\newblock Optimizing transmission infrastructure investments to support line de-energization for mitigating wildfire ignition risk.
\newblock \emph{IREP Symposium}, 2022.

\bibitem[Lummus et~al.(2001)Lummus, Krumwiede, and Vokurka]{lummus2001relationship}
R.~R. Lummus, D.~W. Krumwiede, and R.~J. Vokurka.
\newblock The relationship of logistics to supply chain management: Developing a common industry definition.
\newblock \emph{Industrial Management \& Data Systems}, 101\penalty0 (8):\penalty0 426--432, 2001.

\bibitem[Madhavan et~al.(2020)Madhavan, Kerr, Corcos, and Isaacoff]{madhavan2020toward}
R.~Madhavan, J.~A. Kerr, A.~R. Corcos, and B.~P. Isaacoff.
\newblock Toward trustworthy and responsible artificial intelligence policy development.
\newblock \emph{IEEE Intelligent Systems}, 35\penalty0 (5):\penalty0 103--108, 2020.

\bibitem[Meca and Timmer(2008)]{meca2008supply}
A.~Meca and J.~B. Timmer.
\newblock Supply chain collaboration.
\newblock In \emph{Supply Chain, Theory and Applications}, pages 1--18. I-Tech Education and Publishing, 2008.

\bibitem[Mitchell(2013)]{mitchell2013power}
J.~W. Mitchell.
\newblock Power line failures and catastrophic wildfires under extreme weather conditions.
\newblock \emph{Engineering Failure Analysis}, 35:\penalty0 726--735, 2013.

\bibitem[OECD(2020)]{OECD2020}
OECD, 2020.
\newblock URL \url{https://www.oecd.org/coronavirus/policy-responses/food-supply-chains-and-covid-19-impacts-and-policy-lessons-71b57aea/}.

\bibitem[Palle(2023)]{palle2023evolutionary}
R.~R. Palle.
\newblock Evolutionary optimization techniques in {AI}: Investigating evolutionary optimization techniques and their application in solving optimization problems in {AI}.
\newblock \emph{Journal of Artificial Intelligence Research}, 3\penalty0 (1):\penalty0 1--13, 2023.

\bibitem[Parliament and of~the European~Union(2024)]{eu_ai_act_trilogue_2024}
E.~Parliament and C.~of~the European~Union.
\newblock Regulation {(EU)} 2023/1111 of the {European Parliament and of the Council on Artificial Intelligence}, 2024.
\newblock URL \url{https://artificialintelligenceact.eu/wp-content/uploads/2024/02/AIA-Trilogue-Coreper.pdf}.
\newblock Accessed: 2024-09-12.

\bibitem[PG\&E(2024)]{PGE_undergrounding}
PG\&E.
\newblock System hardening and undergrounding, February 2024.
\newblock URL \url{https://www.pge.com/en/outages-and-safety/safety/community-wildfire-safety-program/system-hardening-and-undergrounding.html}.

\bibitem[Phillipson et~al.(2020)Phillipson, Gorton, Turner, Shucksmith, Aitken-McDermott, Areal, Cowie, Hubbard, Maioli, McAreavey, et~al.]{phillipson2020covid}
J.~Phillipson, M.~Gorton, R.~Turner, M.~Shucksmith, K.~Aitken-McDermott, F.~Areal, P.~Cowie, C.~Hubbard, S.~Maioli, R.~McAreavey, et~al.
\newblock The {COVID-19} pandemic and its implications for rural economies.
\newblock \emph{Sustainability}, 12\penalty0 (10):\penalty0 3973, 2020.

\bibitem[Piansky et~al.(2025)Piansky, Taylor, Rhodes, Molzahn, Roald, and Watson]{piansky2025hicss}
R.~Piansky, S.~Taylor, N.~Rhodes, D.~K. Molzahn, L.~A. Roald, and J.-P. Watson.
\newblock Quantifying metrics for wildfire ignition risk from geographic data in power shutoff decision-making.
\newblock \emph{{\rm To appear in} 58th Hawaii International Conference on System Sciences (HICSS)}, January 2025.

\bibitem[Pollack et~al.(2024)Pollack, Piansky, Gupta, Kody, and Molzahn]{pollack2024equitably}
M.~Pollack, R.~Piansky, S.~Gupta, A.~Kody, and D.~Molzahn.
\newblock Equitably allocating wildfire resilience investments for power grids: The curse of aggregation and vulnerability indices.
\newblock \emph{arXiv preprint arXiv:2404.11520}, 2024.

\bibitem[Radu(2021)]{radu2021steering}
R.~Radu.
\newblock Steering the governance of artificial intelligence: National strategies in perspective.
\newblock \emph{Policy and Society}, 40\penalty0 (2):\penalty0 178--193, 2021.

\bibitem[Rhodes et~al.(2023)Rhodes, Coffrin, and Roald]{rhodes2023security}
N.~Rhodes, C.~Coffrin, and L.~Roald.
\newblock Security constrained optimal power shutoff.
\newblock \emph{arXiv preprint arXiv:2304.13778}, 2023.

\bibitem[Saif and Elhedhli(2016)]{SAIF2016274}
A.~Saif and S.~Elhedhli.
\newblock Cold supply chain design with environmental considerations: A simulation-optimization approach.
\newblock \emph{European Journal of Operational Research}, 251\penalty0 (1):\penalty0 274--287, 2016.

\bibitem[Schewe(2007)]{schewe2007grid}
P.~F. Schewe.
\newblock \emph{The grid: A journey through the heart of our electrified world}.
\newblock National Academies Press, 2007.

\bibitem[St-Arnaud et~al.(2022)St-Arnaud, Carvalho, and Farnadi]{st2022adaptation}
W.~St-Arnaud, M.~Carvalho, and G.~Farnadi.
\newblock Adaptation, comparison and practical implementation of fairness schemes in kidney exchange programs.
\newblock \emph{arXiv preprint arXiv:2207.00241}, 2022.

\bibitem[Taha(2013)]{taha2013operations}
H.~A. Taha.
\newblock \emph{Operations Research: An Introduction}.
\newblock Pearson Education India, 2013.

\bibitem[Taylor and Roald(2022)]{taylor2021wildfire}
S.~Taylor and L.~A. Roald.
\newblock A framework for risk assessment and optimal line upgrade selection to mitigate wildfire risk.
\newblock \emph{Electric Power Systems Research}, 213:\penalty0 108592, 2022.
\newblock Presented at \emph{22nd Power Systems Computation Conference (PSCC)}, June 2022.

\bibitem[Thirunavukkarasu et~al.(2023)Thirunavukkarasu, Sawle, and Lala]{thirunavukkarasu2023comprehensive}
M.~Thirunavukkarasu, Y.~Sawle, and H.~Lala.
\newblock A comprehensive review on optimization of hybrid renewable energy systems using various optimization techniques.
\newblock \emph{Renewable and Sustainable Energy Reviews}, 176:\penalty0 113192, 2023.

\bibitem[Valentim et~al.(2019)Valentim, Louren{\c{c}}o, and Antunes]{valentim2019impact}
I.~Valentim, N.~Louren{\c{c}}o, and N.~Antunes.
\newblock The impact of data preparation on the fairness of software systems.
\newblock In \emph{30th IEEE International Symposium on Software Reliability Engineering (ISSRE)}, pages 391--401. IEEE, 2019.

\bibitem[Van~Hentenryck and Dalmeijer(2024)]{van2024ai4opt}
P.~Van~Hentenryck and K.~Dalmeijer.
\newblock {AI4OPT: AI Institute for Advances in Optimization}.
\newblock \emph{AI Magazine}, 45\penalty0 (1):\penalty0 42--47, 2024.

\bibitem[Veale et~al.(2018)Veale, Van~Kleek, and Binns]{veale2018fairness}
M.~Veale, M.~Van~Kleek, and R.~Binns.
\newblock Fairness and accountability design needs for algorithmic support in high-stakes public sector decision-making.
\newblock In \emph{Proceedings of the 2018 Chi Conference on Human Factors in Computing Systems}, 2018.

\bibitem[Wang and Zhao(2021)]{wang2021cold}
M.~Wang and L.~Zhao.
\newblock Cold chain investment and pricing decisions in a fresh food supply chain.
\newblock \emph{International Transactions in Operational Research}, 28\penalty0 (2):\penalty0 1074--1097, 2021.

\bibitem[Weaver(2021)]{weaver4federal}
J.~F. Weaver.
\newblock The federal government and trustworthy {AI}.
\newblock \emph{The Journal of Robotics, Artificial Intelligence \& Law}, 4, 2021.

\bibitem[{White House}(2023)]{White_House_2023}
{White House}.
\newblock \emph{{Justice40: A whole-of-government initiative}}.
\newblock "The White House", Apr 2023.
\newblock URL \url{https://www.whitehouse.gov/environmentaljustice/justice40/}.

\bibitem[{White House Office of Science and Technology Policy}(2023)]{ai_bill_of_rights}
{White House Office of Science and Technology Policy}.
\newblock Ai bill of rights, 2023.
\newblock URL \url{https://www.whitehouse.gov/ostp/ai-bill-of-rights/#discrimination}.
\newblock Accessed: 2024-09-12.

\bibitem[Whittlestone and Clarke(2022)]{whittlestone2022ai}
J.~Whittlestone and S.~Clarke.
\newblock {AI} challenges for society and ethics.
\newblock In \emph{The Oxford Handbook of AI Governance}. Oxford University Press, 2022.

\bibitem[Zhou et~al.(2020)Zhou, Chen, and Holzinger]{zhou2020towards}
J.~Zhou, F.~Chen, and A.~Holzinger.
\newblock Towards explainability for {AI} fairness.
\newblock In \emph{International Workshop on Extending Explainable AI Beyond Deep Models and Classifiers}, pages 375--386. Springer, 2020.

\end{thebibliography}

\newpage
\appendix
\section*{Appendix}
\section{Technical and Contextual Details for Collaborative Decentralized Cold Supply Chains} \label{Apen:SCL}

\subsection{Supply Chain and Logistics Background}

Supply chain and logistics networks play a pivotal role in today’s global economy. The distinction between supply chain and logistics is important to understand:
\begin{itemize}
    \item \textbf{Logistics}: Defined as the process of “relating essentially to the movement and transmittal of goods, services, and information” \citep{lummus2001relationship}.
    \item \textbf{Supply Chain}: A broader concept that "links all of the partners in the chain including departments within an organization and external partners like suppliers, carriers, and third-party companies" \citep{lummus2001relationship}.
\end{itemize}

For efficient supply chain and logistics design, it’s crucial to view the entire process as a single, interconnected system to improve operations and competitiveness. These networks demonstrated significant vulnerabilities during the COVID-19 pandemic, exposing weaknesses in global logistics, especially with regards to food supply chains.

\subsection{Cold Food Supply Chains Overview}

Cold food supply chains are a subset of logistics dedicated to the transportation and storage of temperature-sensitive goods. These include perishable food products like dairy, meat, and produce, which require constant temperature control to maintain quality and safety \citep{awad2021vehicle}. The challenges in cold supply chains are unique and involve:
\begin{itemize}
    \item \textbf{Energy-Intensive Equipment}: Temperature control equipment, such as refrigerated trucks, demands significant energy consumption, leading to higher costs \citep{SAIF2016274}.
    \item \textbf{Specialized Logistics Capabilities}: Rural areas, in particular, often lack the necessary infrastructure for efficient cold chain management, which leads to higher spoilage rates and transportation costs.
\end{itemize}

Managing cold supply chains in rural areas is particularly challenging due to underdeveloped infrastructure, leading to logistical hurdles for small-scale farms.

\subsection{Collaborative and Non-Collaborative Supply Chains}

\begin{itemize}
    \item \textbf{Collaborative Supply Chains}: Multiple organizations form a coalition, pooling resources to optimize and synchronize operations. The focus is on improving efficiency, reducing costs, and enhancing service quality by sharing inventory, transportation, and demand forecasting among members \citep{GUAJARDO2015147}.
    \item \textbf{Non-Collaborative Supply Chains}: In contrast, each organization operates independently, optimizing its own processes without consideration for the broader network \citep{meca2008supply}. This results in inefficiencies and higher costs when organizations act in isolation.
\end{itemize}

A collaborative supply chain is especially beneficial for smaller entities that may not have the resources to tackle challenges independently. Decentralization of decision-making and operations allows these organizations to respond more flexibly to local conditions \citep{JUNG2008348}.

\subsection{Mathematical Formulation of Collaborative Decentralized Cold Food Supply Chain}

The collaborative decentralized cold food supply chain problem involves multiple farms and distribution centers working together to ensure that the maximum number of farms and communities are served, while ensuring fairness and equity in resource allocation. The objective is to maximize the coverage of farms and communities, ensuring that resources are distributed equitably, particularly to underrepresented populations.

The mathematical formulation can be expressed as:

\begin{subequations}
\begin{align}
\text{Maximize} \quad & \sum_{i \in \mathcal{F}} \sum_{j \in \mathcal{D}} X_{ij} + \sum_{k \in \mathcal{C}} Y_k \label{eq:CDCFSC:obj} \\
\text{Subject to:} \quad & \sum_{i \in \mathcal{F}} X_{ij} \leq D_j, \quad \forall j \in \mathcal{D} \label{eq:CDCFSC:dist}\\
& \sum_{j \in \mathcal{D}} X_{ij} \leq F_i, \quad \forall i \in \mathcal{F} \label{eq:CDCFSC:farm}\\
& \sum_{k \in \mathcal{C}} Y_k \leq M, \quad \forall k \in \mathcal{C} \label{eq:CDCFSC:cap}\\
& X_{ij}, Y_k \in \{0, 1\}, \quad \forall i,j,k \label{eq:CDCFSC:dom}\\
\end{align}
\end{subequations}

In this formulation, $\textcolor{black}{\mathcal{F}}$ represents the set of farms, $\textcolor{black}{\mathcal{D}}$ represents the set of distribution centers, and $\textcolor{black}{\mathcal{C}}$ represents the set of communities. The binary decision variable $X_{ij}$ indicates whether farm $i$ is connected to distribution center $j$ (1 if covered, 0 otherwise), and $Y_k$ is a binary decision variable that represents whether community $k$ is served (1 if served, 0 otherwise). $D_j$ refers to the demand at distribution center $j$, while $F_i$ represents the supply capacity from farm $i$. The total storage or transportation capacity is denoted by $M$. The objective function \eqref{eq:CDCFSC:obj} aims to maximize the total number of farms and communities covered by the supply chain, subject to constraints \eqref{eq:CDCFSC:farm} that respect the supply limits of farms, constraints \eqref{eq:CDCFSC:dist} that represent the demands of distribution centers, and constraints \eqref{eq:CDCFSC:cap} that adhere to storage or transportation capacity limits. The formulation prioritizes maximizing coverage across both farms and communities, ensuring that underrepresented populations benefit equitably from the collaborative cold food supply chain. Additional fairness constraints can be incorporated to ensure that smaller farms or more vulnerable communities are not excluded from resource allocation.

\section{Technical and Conceptual Details for the Power Systems Case Study} \label{Appen:PwrSys}

\subsection{Technical Details on Power Grid Representation and OTS Problem} \label{Apen:OTS}

Power grids are commonly modeled as a network of buses (vertices) connected by power lines (edges). These networks are critical for understanding how electricity is distributed and managed across large geographic areas. The Optimal Transmission Switching (OTS) problem is a subset of Network Topology Optimization (NTO), which focuses on determining the most efficient configuration of the power grid by switching certain transmission lines on or off. The OTS problem becomes particularly important in scenarios with high wildfire ignition risk, as switching off certain lines can reduce the likelihood of a wildfire being started by electrical infrastructure.

During periods of elevated wildfire risk, utilities employ Public Safety Power Shutoff (PSPS) events, which involve de-energizing lines to prevent fires. However, this can also result in load shedding, where customers experience power outages. The challenge in the OTS-PSPS problem is to find an optimal balance between reducing wildfire risk and minimizing the amount of power that is not served (i.e., minimizing load shed)~\citep{taylor2021wildfire, rhodes2023security, kody2022optimizing}.

\subsection{Mathematical Formulation of OTS Problem} \label{Apen:MathOTS}

The general form of the Optimal Transmission Switching (OTS) problem can be expressed as:

\begin{subequations}
\begin{align}
\text{Minimize} \quad & f(x, u) 
\label{eq:OTS:obj}\\
\text{Subject to:} \quad & g(x, u) \leq 0 
\label{eq:OTS:ineq} \\
& h(x, u) = 0 \label{eq:OTS:eq}\\
& x \in X, \, u \in U \label{eq:OTS:domain}
\end{align}
\end{subequations}

Here, $x$ represents the state of the power system (including power flows and voltage levels), while $u$ represents the decision variables for transmission switching. The objective \eqref{eq:OTS:obj} minimizes total operational costs while considering the load shed (power not served). Constraints \eqref{eq:OTS:ineq}--\eqref{eq:OTS:domain} represent the physical limits of the grid and the operational constraints during wildfire risk, such as power flow limits, voltage limits, and the risk of wildfire ignition.

\subsection{Utility Investments in Line Undergrounding}

Several utility companies have made significant commitments to undergrounding their existing power infrastructure to mitigate the risk of wildfires. Pacific Gas and Electric (PG\&E), for example, has pledged to spend billions of dollars on this initiative over the next several years. Underground power lines offer a solution to the wildfire ignition risk posed by above-ground lines, as they can continue to deliver power without contributing to the ignition of wildfires~\citep{PGE_undergrounding}. 

While these investments are necessary to reduce risk, they raise critical questions about equitable resource allocation. Marginalized communities, who often face disproportionate impacts from power outages and wildfires, must be included in the benefits of these infrastructure upgrades. Federal guidelines, such as those from the Justice40 initiative, highlight the importance of ensuring that climate-related infrastructure investments prioritize vulnerable populations.

\subsection{Justice40 Initiative Overview}

The Justice40 initiative, enacted through Executive Order by the Biden-Harris administration, aims to ensure that 40\% of the overall benefits of certain federal investments flow to disadvantaged communities that have been historically marginalized and disproportionately affected by environmental hazards. These communities are identified using a variety of factors, including economic, environmental, and public health data detailed in the Climate \& Economic Justice Screening Tool (CEJST). The initiative applies to projects focused on climate-related infrastructure, including clean energy, affordable housing, and water and wastewater infrastructure improvements~\citep{White_House_2023}.

\subsection{Data Sources and Integration Methods}

In Case Study 1, discussed in Section \ref{PwrCase}, we used three key data sources to assess wildfire ignition risk and its effects on marginalized communities: the Justice40 designated census tracts, the CDC Social Vulnerability Index (SVI), and the USGS Wildland Fire Potential Index (WFPI). Here is a summary of how each dataset was integrated into the model:

\begin{itemize}
    \item \textbf{Justice40-designated Census Tracts}: These tracts are defined by the Justice40 initiative as areas where 40\% of the benefits of federally funded climate-related infrastructure must flow. We mapped the census tracts to buses in the transmission grid using a three-step proximity method:
    \begin{enumerate}
        \item Find the closest bus to each census tract center.
        \item Iterate over buses with load and assign census tracts to the nearest bus.
        \item Distribute population evenly among all buses within the calculated proximity radius.
    \end{enumerate}

    \item \textbf{CDC Social Vulnerability Index (SVI)}: The SVI provides vulnerability metrics across four subcategories: socioeconomic status, household characteristics, racial and ethnic minority status, and housing type \& transportation. We aggregated these submetrics into a single vulnerability score for each census tract. Alternative aggregation methods, such as weighting individual submetrics more heavily, could yield different outcomes. The SVI scores were then correlated to buses on the grid in the method discussed above.

    \item \textbf{USGS Wildland Fire Potential Index (WFPI)}: The WFPI offers daily, unitless ignition risk values at a 1~km by 1~km grid resolution. We calculated ignition risk for each transmission line by summing the risk values along the line, considering any values above one standard deviation from the historical mean as high risk. This allows us to create a "high-risk" line integral to use in our OTS formulation. \cite{piansky2025hicss} provides more information on impacts of various risk aggregation methods.

    \item \textbf{Population Allocation to Transmission Buses}: The method used in this case study splits population proportionally across buses based on the distance from census tracts, as discussed above, but alternative strategies could alter the outcomes of the model.
\end{itemize}

Each data source was carefully pre-processed and normalized to ensure consistency. The specific method of data correlation and the choice of vulnerability index or wildfire risk data can affect the model’s results, leading to different interpretations of how resources should be allocated to mitigate wildfire risk.


\end{document}